\documentclass[letterpaper]{llncs}
\usepackage{times}
\usepackage{helvet}
\usepackage{courier}
\usepackage{epic}
\usepackage{graphicx}
\usepackage{latexsym}
\usepackage{verbatim}
\usepackage{xspace}
\usepackage[linesnumbered, vlined,ruled]{algorithm2e}
\usepackage{amsfonts}
\usepackage{amsmath}
\usepackage{amssymb}
\begin{document}

\newlength{\halftextwidth}
\setlength{\halftextwidth}{0.47\textwidth}
\def\halffigsize{2.2in}
\def\thirdfigsize{1.5in}
\def\negvspace{0in}
\def\posvspace{0em}

\newcommand{\Ddpv}{\ensuremath{\mathcal D^{dp}_v}}
\newcommand{\Ddp}{\ensuremath{\mathcal D^{dp}}}
\newcommand{\myset}[1]{\ensuremath{\mathcal #1}}
\newcommand{\nina}[1]{\marginpar{\sc nina}\textit{#1}}
\renewcommand{\nina}[1]{#1}

\newtheorem{mydefinition}{Definition}
\newtheorem{myexample}{Example}
\newtheorem{myobservation}{Observation}
\newtheorem{mytheorem}{Theorem}
\newtheorem{mylemma}{Lemma}
\newtheorem{mycor}{Corollary}
\newenvironment{runexample}{{\bf Running example:} \it}{\rm}
\newcommand{\myprroof}{\noindent {\bf Proof:\ \ }}
\newcommand{\myqed}{\mbox{$\Box$}}

\newcommand{\alldifferent}{\mbox{\sc AllDifferent}\xspace}
\newcommand{\alldiff}{\mbox{\sc AllDifferent}\xspace}
\newcommand{\alldiffprec}{\mbox{\sc AllDiffPrec}\xspace}
\newcommand{\cumulative}{\mbox{\sc Cumulative}\xspace}
\newcommand{\alldiffprecbf}{\mbox{\sc AllDiffPrec}\xspace}

\newcommand{\mymax}{\mbox{\rm max}}
\newcommand{\mymin}{\mbox{\rm min}}
\newcommand{\mylog}{\mbox{\rm log}}

\newcommand{\todo}[1]{{\tt (... #1 ...)}}

\title{The AllDifferent Constraint with Precedences\thanks{
Supported by the Australian
Government's  Department of Broadband, Communications and the Digital Economy
and the
ARC.}}
\author{
Christian Bessiere\inst{1},
Nina Narodytska\inst{2},
Claude-Guy Quimper\inst{3}
\and
Toby Walsh\inst{2}}
\institute{CNRS/LIRMM, Montpellier, email: bessiere@lirmm.fr
\and NICTA and University of NSW,
Sydney, email:
\{nina.narodytska,toby.walsh\}@nicta.com.au
\and
Universit\'{e} Laval,
Qu\'{e}bec, email: claude-guy.quimper@ift.ulaval.ca
}

\maketitle
\begin{abstract}
We propose \alldiffprec, a new global constraint that combines together an \alldiff
constraint with precedence constraints that strictly order given pairs
of variables.
We identify a number of applications for this global constraint including
instruction scheduling and symmetry breaking. We give an efficient
propagation algorithm that enforces bounds consistency on this global
constraint. We
show how to implement this propagator using a decomposition that
extends the bounds consistency enforcing decomposition proposed for the
\alldiff constraint. Finally, we prove that enforcing domain
consistency on this global constraint is NP-hard in general.
\end{abstract}
\sloppy

\section{Introduction}

One of the important features of constraint programming
are global constraints. These capture
common modelling patterns (e.g. ``these jobs need to be
processed on the same machine so must take place
at different times'').
In addition, efficient propagation algorithms
are associated with global constraints
for pruning the search space (e.g.
``these 5 jobs have only 4 time slots between
them so, by a pigeonhole argument,
the problem is infeasible'').
One of the oldest and most useful global constraints is the \alldifferent
constraint~\cite{alice}. This specifies that
a set of variables takes all different values.
Several algorithms have been proposed for
propagating 
this constraint
(e.g. \cite{regin1,Leconte,puget98,mtcp2000,lopez1}).
Such propagators can have
a significant impact on our ability
to solve problems (see, for instance, \cite{swijcai99}).
It is not hard to provide pathological problems
on which some of these propagation algorithms provide
exponential savings. 
A number of hybrid frameworks have been proposed to
combine the benefits of such propagation algorithms
and OR methods like integer linear
programming (see, for instance, \cite{mortinforms2002}).
In addition, the convex hull of a number of global
constraints has been studied in detail (see, for instance,
\cite{wyinforms2001}).

In this paper, we consider a modelling pattern
\cite{cp2003} that
occurs in many problems involving \alldiff constraints.
In addition to the constraint that no pair of variables
can take the same value, we may also have a constraint that
certain pairs of variables are ordered 
(e.g.
``these two jobs need to be
processed on the same machine so must take place
at different times, but the first job must be processed
before the second'').
\nina{
We propose a new
global constraint, \alldiffprec that captures this pattern.
This 
global constraint is a specialization of the general
framework that combines several \cumulative  and precedence
constraints~\cite{Beldiceanu96,Simonis99}.
Reasoning about such combinations of global constraints
may achieve additional pruning. 
In this work} we propose an efficient propagation algorithm for
the \alldiffprec constraint. However, 
we also prove that 
propagating the constraint {\em completely}
is computationally intractable.

\section{Formal background}

A constraint satisfaction problem (CSP) consists of a set of
variables, each with a domain of possible values, and a set of
constraints specifying allowed values for
subsets of variables.
A solution is an assignment of values to the variables
satisfying the constraints.
We write $\myset{D}(X)$
for the domain of the variable $X$.
Domains can be ordered (e.g. integers).
In this case, we write $min(X)$ and $max(X)$ for the minimum and
maximum elements in $\myset{D}(X)$.
The {\em scope} of a constraint
is the set of variables to which it is applied.
A \emph{global constraint} is one in which the number of variables
is not fixed. 
For instance,
the global constraint
$\alldiff([X_1,\ldots,X_n])$ ensures $X_i \neq X_j$ for
$1 \leq i < j \leq n$.
By comparison, the binary constraint, $X_i \neq X_j$
is not global.

When solving a CSP, we often use
propagation algorithms to prune the search
space by enforcing properties like domain, bounds or range
consistency.
A \emph{support} on a constraint $C$ is an assignment of all variables
in the scope of $C$ to values in their domain such  that $C$ is satisfied. 
A variable-value $X_i = v$ is \emph{consistent} on $C$ iff it belongs to a support
of $C$. 
A constraint $C$ is \emph{domain consistent} (\emph{DC})
iff every value in the domain of every variable in the scope of $C$ is
consistent on $C$. 
A \emph{bound support} on $C$  is an assignment of all variables in
the scope of $C$ to values  between their minimum  and maximum values
(respectively called lower and upper bound) 
such that $C$ is satisfied. 
A variable-value $X_i = v$ is \emph{bounds consistent} on $C$ iff it
belongs to a bound support of $C$. 
A constraint $C$ is   \emph{bounds consistent} (\emph{BC})
iff
the  lower and upper bounds of every variable in the scope of $C$ are
bounds consistent on $C$. 
Range consistency
is stronger than  $BC$ but is weaker than $DC$.
A constraint $C$ is   \emph{range consistent} (\emph{RC}) iff 
iff every value in the domain of every variable in the scope of $C$ is
bounds consistent on $C$. 
A CSP is $DC$/$RC$/$BC$ iff each constraint is $DC$/$RC$/$BC$.
Generic algorithms exists for enforcing
such local consistency properties. For
global constraints like \alldiff, specialized
methods have also been developed which
offer computational efficiencies.
\nina{For example, a bounds consistency propagator for \alldiff
is based on the notion of Hall interval. 
A Hall interval is an interval of
$h$ domain values that completely contains
the domains of $h$ variables.  Clearly,
variables whose domains are contained within
the Hall interval consume all the values
in the Hall interval, whilst any other variables 
must find their support outside the Hall interval.
}

\newcommand{\tighter}{\mbox{$\preceq$}}
\newcommand{\stighter}{\mbox{$\prec$}}
\newcommand{\incomparable}{\mbox{$\bowtie$}}
\newcommand{\equivalent}{\mbox{$\equiv$}}

We will compare local consistency properties applied to 
logically equivalent constraints.
As in \cite{debruyne1}, we say that
a local consistency property $\Phi$ on the set of 
constraints $S$ is {\em stronger} than
$\Psi$ on the logically equivalent set $T$
iff, given any domains, $\Phi$ removes all values  $\Psi$ removes, and
sometimes more.
For example, 
domain consistency on $\alldiff([X_1,\ldots,X_n])$
is stronger than domain consistency on $\{ X_i \neq X_j \ \mid 1
\leq i < j \leq n\}$. In other words, decomposition of the
global \alldiff constraint into binary
not-equals constraints hinders propagation.

\section{Some examples}

To motivate the introduction of this
global constraint, we give some examples of models
where we have one or more sets of
variables which take all-different values, as well as
certain pairs of these variables which are ordered.

\subsection{Exam time-tabling}

Suppose we are time-tabling exams.
A straight forward model has variables
for exams, and values which are the possible
times for these exams. In such a model,
we may have temporal precedences (e.g. part 1 of
the physics exam must be before part 2) as well
as \alldiff\ constraints on those sets of exams with
students in common (e.g.
all physics, maths, and chemistry exams
must occur at different times since there are
students that need to sit all three exams).

\subsection{Scheduling}

Suppose we are
scheduling
a single machine with unit-time tasks, subject to precedence
constraints and release and due times \cite{gjstsiam81}.
A straight forward model has variables
for the tasks, and values which are the possible
times that we execute each task. In such a model,
we have an $\alldiffprec$ constraint on variables
whose domains are the appropriate intervals.
For example, consider scheduling instructions in a block
(a straight-line sequence of code with a single
entry and exit point) on one processor where all instructions
take the same time to execute.
Such a schedule is subject to a number of different types
of precedence constraints.
For instance, instruction {$\cal A$}
must execute before {$\cal B$} if:
\begin{description}
\item[Read-after-write dependency:]
{$\cal B$} reads a register written by {$\cal A$};
\item[Write-after-write dependency:]
{$\cal B$} writes a register also written by {$\cal A$};
\item[Write-after-read dependency:]
{$\cal B$} writes a register that {$\cal A$} reads.
\end{description}
Such dependencies give rise to precedence
constraints between the instructions.

\subsection{Breaking value symmetry}

Many constraint models contain value symmetry.
Puget has proposed a
general method for breaking any
number of value symmetries in polynomial time \cite{pcp05,pcpl2007}.
This method introduces
variables $Z_j$ to represent the index
of the first occurrence of each value:
\begin{eqnarray*}
X_i = j   \Rightarrow  Z_j \leq i, & \ \ \ \ \ & 
Z_j = i \Rightarrow X_i=j
\end{eqnarray*}
Value symmetry on the $X_i$ is transformed
into variable symmetry on the $Z_j$.
This variable symmetry is 
easy to break. We simply need to post
precedence constraints on the $Z_j$.
Depending on the value symmetry, we
need different precedence constraints.

Consider, for example, finding a graceful labelling of
a graph. A graceful labelling is a labelling of the vertices
of a graph with distinct integers 0 to $e$ such that the $e$
edges (which are labelled with the absolute
differences of the labels of the two connected
vertices) are also distinct. Graceful labellings have
applications in radio astronomy, communication networks,
X-ray crystallography, coding theory and elsewhere.
Here is the graceful labelling of the graph
$K_3 \times P_2$:
\begin{center}
\scalebox{0.7}{\includegraphics{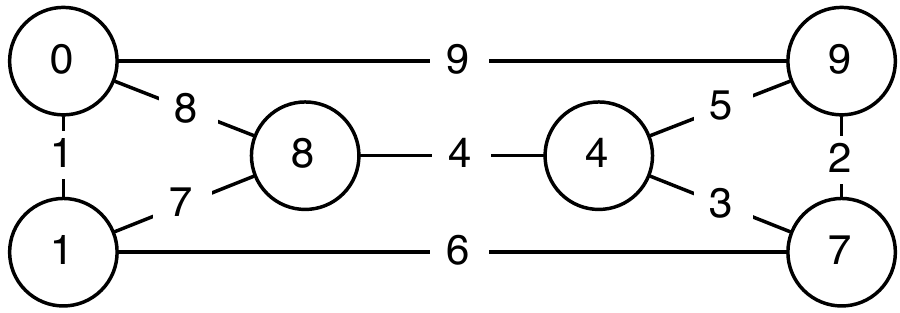}}
\end{center}
A straight forward model for graceful labelling
a graph has variables for
the vertex labels, and values which are integers
0 to $e$. This model has a simple value symmetry
as we can map every value $i$ onto $e-i$.
In \cite{pcpl2007}, Puget breaks this value symmetry
for $K_3 \times P_2$ with
the following ordering constraints:
$$ Z_0 < Z_1, \ Z_0 < Z_3, \ Z_0 < Z_4, \ Z_0 < Z_5, \  Z_1 < Z_2
$$
Note that all the $Z_j$ take different values as each
integer first occurs in the graph at a different index.
Hence, we have a sequence of variables on
which there is both
an \alldiff and precedence
constraints.

\section{\alldiffprec}
\label{s:alldiffprec}

Motivated by such examples, we propose the global constraint:
$$\alldiffprec([X_1,\ldots,X_n],E)$$
Where $E$ is a set containing pairs of variable indices.
This ensures $X_i \neq X_j$ for any $1\leq i < j \leq n$
and $X_j < X_k$ for any $(j,k) \in E$.
Without loss of generality, we assume that $E$ does not
contain cycles. If it does, the constraint is trivially unsatisfiable.
It is not hard to see that decomposition of
this global constraint into separate \alldiff\ and binary
ordering constraints can hinder propagation.
\begin{mylemma}
Domain consistency on the constraint $\alldiffprec ([X_1,\ldots,X_n],E)$
is stronger
than domain consistency on the decomposition into
$\alldiff ([X_1,$ $\ldots,X_n])$ and
the binary ordering constraints,
$X_i < X_j$ for $(i,j) \in E$.
Bounds
consistency on $\alldiffprec ([X_1,\ldots,X_n],E)$
is stronger than bounds consistency on the decomposition,
whilst range
consistency on $\alldiffprec ([X_1,\ldots,X_n],E)$
is stronger than range consistency on the decomposition.
\end{mylemma}
\myprroof
Consider $\alldiffprec ([X_1,X_2,X_3],\{(1,3),(2,3)\})$
with $\myset{D}(X_1)=\myset{D}(X_2)= \{1,2,3\}$ and
$\myset{D}(X_3)= \{2,3,4\}$. Then
the decomposition into $\alldiff([X_1,X_2,X_3])$ and
the binary ordering constraints,
$X_1 < X_3$, and $X_2 < X_3$
is domain consistent. Hence, it is also range and
bounds consistent.
However, enforcing bounds consistency
directly on the global $\alldiffprec$ constraint will prune
2 from the domain of $X_3$
since this assignment has no bound support.
Similarly, enforcing range or domain consistency
will prune 2 from the domain of $X_3$.
\myqed

A simple greedy method will find a bound support
for the \alldiffprec constraint. This method
is an adaptation of the greedy method to build
a bound support of the \alldiff constraint.
For simplicity, we suppose that $E$ contains the
transitive closure of the precedence constraints.
In fact, this step is not required but makes our
argument easier.
First, we need to preprocess variables domains so that they
respect the precedence constraints $X_i  < X_j$, $(i,j) \in E$:
$\min(X_i) < \min(X_j)$ and $\max(X_i) < \max(X_j)$. However, we notice that it is sufficient
to enforce a weaker condition on bounds of variables $X_i$ and $X_j$ such that
$\min(X_i) \leq \min(X_j)$ and  $\max(X_i) \leq  \max(X_j)$.
If these conditions on variables domains are satisfied then
we say that domains are \emph{preprocessed}.
Second, we construct a satisfying assignment as follows.
We process all
values in the increasing order. When processing a value $v$, we assign $v$ to the
variable with the smallest upper bound, $u$ that has not yet been assigned and
that contains $v$ in its domain. Suppose, there exists a set of variables that
have the upper bound $u$, so that $X' = \{X_i \mid \myset{D}(X_i) = [v,
u]\}$.
To construct a solution for \alldiff, we would break these ties
arbitrarily.
In this case, however, we select a variable that is
not successor of any variable in the set $X'$. Such a variable always exists, as the transitive closure of the precedence
graph does not contain cycles. By the correctness of the original algorithm the
resulting assignment is a solution.
In addition to satisfying the \alldiff constraint, this solution also
satisfies the precedence constraints. Indeed, for the constraint $X_i < X_j$, the
upper bound of $\myset{D}(X_i)$ is necessarily smaller than or equal to the upper bound
of $\myset{D}(X_j)$. In the case of equality, we tie break in favor of $X_i$. Therefore,
a value is assigned to $X_i$ before a value gets assigned to 
$X_j$ . Since we process values in increasing order, we obtain $X_i <
X_j$ as required.

\begin{myexample}
\label{exm:greedy}
Consider $\alldiffprec ([X_1,X_2,X_3,X_4],\{(1,3),(2,3),(1,4),(2,4)\})$
with $\myset{D}(X_1)=\myset{D}(X_2)= \{1,2,3,4,5\}$,
$\myset{D}(X_3)= \{1,2,3\}$ and $ \myset{D}(X_4)= \{2,3,4\}$.
First, we preprocess domains to ensure that
$\min(X_i) \leq \min(X_j)$ and  $\max(X_i) \leq  \max(X_j)$, $ i \in \{1,2\}$, $j \in \{3,4\}$.
This gives
$\myset{D}(X_1)=\myset{D}(X_2)= \myset{D}(X_3) = \{1,2,3\}$,
$\myset{D}(X_4)= \{2,3,4\}$. As in the greedy algorithm,
we consider the first value $1$. This value is contained in domains
of variables $X_1$, $X_2$ and $X_3$. As $\max(X_1)=\max(X_2)=\max(X_3) = 3$, by tie breaking we
select variables that are not successors of any other variables among variables $\{X_1,X_2,X_3\}$.
There are two such variables: $X_1$ and $X_2$.
We break this tie arbitrarily and set $X_1$ to 1. The new domains
are $\myset{D}(X_1)= 1$, $\myset{D}(X_2)= \myset{D}(X_3) = \{2,3\}$,
$\myset{D}(X_4)= \{2,3,4\}$.
The next value we consider is $2$. Again, there exist
two variables that contain this value, and they have the same upper bounds.
By tie-breaking, we select $X_2$. 
Finally, we assign $X_3$ and $X_4$ to 3 and 4 respectively.
\end{myexample}

We can design a filtering algorithm based on this satisfiability
test. By successively reducing a variable domain in halves with
a binary search we can filter the lower and upper bounds of a variable domain
with $O(\mylog d)$ tests where $d$ is the cardinality of the domain.
\nina{Consider, for example, a variable $X$
with the domain $D(X) = [l,u]$. We are looking for a support
for $\min(X)$. At the first step we temporally
fix the domain of X to  the first half so that $D(X)= [l, (u-l)/2]$
and run the bounds disentailment
detection algorithm. If this algorithm fails, we halved the search
and repeat with the other half. If this algorithm does not fail,
we know that there is a value in $[l, (u-l)/2]$ that has a bounds
support. Hence, we continue with the binary search within this half.}
As
each test takes $O(n)$ time and there are $n$ variables to
prune, the total running time is $O(n^2 \mylog
d)$. In the rest of this paper, we improve on this using
sophisticated algorithmic ideas.

\section{Bounds consistency}

We present an algorithm that enforces bounds consistency on the
\alldiffprec constraint.
First, we consider an assignment $X_i = v$ and a partial filtering  that
this assignment causes. We call this filtering \emph{direct pruning}
caused by the assignment $X_i = v$ or, in short, direct pruning of $X_i = v$.
Informally, direct pruning works as follows. If $X_i$ takes $v$ then the value $v$
becomes unavailable for the other variables due to the $\alldiff$ constraint.
Hence, we remove $v$ from the domains of variables that have $v$ as their lower
bound or upper bound. Due to precedence constraints, we
increase the lower bounds of successors of $X_i$ to $v + 1$ and decrease the upper
bounds of predecessors of $X_i$ to $v-1$.
Note that direct pruning \emph{does not enforce} bounds consistency on either $\alldiffprec$
or the single $\alldiff$ constraint. However, direct pruning is
sufficient \nina{to detect bounds inconsistency as we show below.}
%
%

Let $P(i)$ and $S(i)$ be the sets of variables  that precede and succeed $X_i$, respectively.
We denote the domains obtained after  direct pruning of $X_i = v$
as $\Ddpv(X_1),\ldots, \Ddpv(X_n)$, so that for all $j =1,\ldots,n$:

\begin{eqnarray}
\Ddpv(X_j) = \myset{D}(X_j) \setminus \{ v\} & \mathrm{if}&  j \neq i, v \in \{\min(X_j),  \max(X_j)\} \label{eq:dom_pruned_1}\\
\Ddpv(X_j) = v &\mathrm{if}&  j = i,\\
\Ddpv(X_j) = \myset{D}(X_j) \setminus  [v, max(X_j)] &\mathrm{if}&  j \in P(i),\\
\Ddpv(X_j) = \myset{D}(X_j) \setminus  [min(X_j), v] &\mathrm{if}&  j \in S(i).  \label{eq:dom_pruned_3}
\end{eqnarray}


These bounds could be pruned further but we will first analyze the properties that this simple filtering offers.

\begin{myexample}
\label{ex:pruning}
Consider  $\alldiffprec ([X_1,X_2,X_3],\{(1,2)\})$ constraint
with $\myset{D}(X_1) =  \{1,2\}$, $\myset{D}(X_2)= \{2,3\}$,
$\myset{D}(X_3)= \{1,2,3\}$.
For example, an assignment $X_1 =2$
results in the domains:
$\Ddp_2(X_1) =  \{2\}$, $\Ddp_2(X_2)= \{3\}$ and
$\Ddp_2(X_3)= \{1,2,3\}$.
We point out again that we can continue pruning as values $2$ and $3$
have to be removed from $\Ddp_2(X_3)$. However, direct pruning of $X_1 = 2$
is sufficient for our purpose. Consider another example. An assignment $X_3 =1$
results in the domains:
$\Ddp_3(X_1) =  \{2\}$, $\Ddp_3(X_2)= \{2,3\}$ and
$\Ddp_3(X_3)= \{1\}$.
\end{myexample}

Our algorithm is based on the following lemma.
\begin{mylemma}
\label{l:testing bound}
Let $\alldiff$ and precedence constraints be bounds consistent \nina{over variables $X$},
$X_i = v$, $v \in  \{\min(X_i), \max(X_i)\}$  be an assignment of a variable $X_i$ to its bound and
$\Ddpv(X_1),\ldots, \Ddpv(X_n)$ be the domains after direct pruning of $X_i=v$.
Then, $X_i = v$ is  bounds consistent iff $\alldiff([X_1,\ldots, X_n])$,
where domains of variables $X$ are $\Ddpv(X_1),\ldots, \Ddpv(X_n)$,
has a solution.
\end{mylemma}
\myprroof
Suppose $\alldiff$ and the precedence constraints are bounds consistent.
As precedence constraints are bounds consistent,
we know that for all $(i,j) \in E$, $X_i  < X_j$,
$\min(X_i) < \min(X_j)$ and $\max(X_i) < \max(X_j)$.
Consider direct pruning of $X_i=v$.
 Note, direct pruning of  $X_i = v$ preserves
the property of domains being preprocessed.
The pruning can only create equality
of lower bounds or upper bounds for some precedence constraints.
The assignment $X_3 = 1$  demonstrates this situation in Example~\ref{ex:pruning}.
Direct pruning of $X_3 = 1$ forces lower bounds
of $X_1$ and $X_2$, that are in the precedence relation, to be equal.

As domains $\Ddpv(X_1),\ldots, \Ddpv(X_n)$ are preprocessed,
we know that the greedy algorithm (Section~\ref{s:alldiffprec})
will find a solution of $\alldiff$  on the domains
$\Ddpv(X_1),\ldots, \Ddpv(X_n)$ that  also satisfies the precedence constraints if a solution exists.
This solution is a support for  $X_i = v$.
\qed

Based on Lemma~\ref{l:testing bound} we prove
\nina{that we can enforce bounds consistency on the $\alldiffprec$ constraint
in $O(n^2)$.
However, we start with a simpler and less efficient
algorithm to explain the idea . We show how to improve this algorithm
in the next section. Given Lemma~\ref{l:testing bound},
the most straightforward algorithm to enforce bounds consistency for $X_i=v$
is to assign $X_i$ to $v$, perform the direct pruning, run the greedy algorithm and,
if it fails, prune $v$.  Interestingly enough, to detect bounds disentailment
we do not have to run a greedy algorithm for  each pair $X_i=v$.
If the $\alldiff$ constraint and the precedence constraints
are bounds consistent, we show that it is sufficient to check
that a set of conditions~\eqref{eq:max_set_1}-\eqref{eq:min_sum}
holds for each interval of values.
If these conditions are satisfied then
the pair $X_i=v$ is bounds consistent. Hence, for each pair $X_i=v$, $1 \leq i \leq n$,
$v \in D(X_i)$,
and for each interval we enforce the following conditions}.
We assume that $\cup_{i=1}^n\myset{D}(X_i) = [1,d]$.
For $X_i$,
$1 \leq i \leq n$, $v \in \myset{D}(X_i)$
and for all intervals  $[v,v+k]$ and $[v-p,v]$,
$k \in [\max(X_i) - v+1,d - v]$ and $p \in [v - \min(X_i)+1, v -1]$,
the following conditions have to be satisfied:
\begin{eqnarray}
B^i_{1,{v+k}}  & = & |\{j \in S(i) |  \myset{D}(X_j) \subseteq [1, v + k]\}|  \label{eq:max_set_1}\\
D^i_{v,(v+k)} & = &|\{j \notin S(i) | \myset{D}(X_j) \subseteq [v, v + k]\}| \label{eq:max_set_2} \\
B^i_{1,{v+k}} + D^i_{v,(v+k)} &\leq& k  \label{eq:max_sum}\\
B^i_{{v-p},d}  & = & |\{j \in P(i) |  \myset{D}(X_j) \subseteq [v - p, d]\}| \label{eq:min_set_1} \\
D^i_{v-p,v} & = &|\{j \notin P(i) | \myset{D}(X_j) \subseteq [v - p, v]\}| \label{eq:min_set_2} \\
B^i_{v-p,d} + D^i_{v-p,v} &\leq& p \label{eq:min_sum}
\end{eqnarray}

\nina{Note that we actually do not have to consider all possible
intervals.
For every variable-value pair $X_i=v$ we consider all intervals
$[v,u]$, $ u \in [\max(X_i)+1, d]$ and all intervals $[l,v]$, $l
\in[1,\min(X_i)-1]$.
The parameter $k$ ($p$) is used to slide between intervals $[v,u]$, $ u \in [\max(X_i)+1, d]$ 
$([l,v]$, $l \in [1,\min(X_i)-1])$.
Equations~\eqref{eq:max_set_1}--\eqref{eq:max_sum} make sure that
the number of variables that fall into an interval $[v,u]$,
after the assignment $X_i$ to $v$, is less than or equal to the length
of the interval minus 1.
Symmetrically, Equations~\eqref{eq:min_set_1}--\eqref{eq:min_sum} ensure that
the same condition is satisfied for all intervals $[l,v]$.
If there exists an interval $[v,u]$($[l,v]$) that violates the
condition for a pair $X_i=v$ then this interval is removed from
$D(X_i)$.}
\begin{myexample}
\label{exm:faster_bc}
Consider $\alldiffprec ([X_1,X_2,X_3,X_4,X_5],\{(1,2),(1,3)\})$.
Domains of the variables are $D(X_1) = [1,5]$, $D(X_2)= D(X_3) = [2,6]$
and $D(X_4)=D(X_5)=[3,6]$.
Consider a variable-value pair $X_1=3$. By the direct pruning
we get the following domains:
$\Ddp_3(X_1) = 3$,
$\Ddp_3(X_2) = [4,6]$,
$\Ddp_3(X_3) = [4,6]$,
$\Ddp_3(X_4) = [4,6]$ and
$\Ddp_3(X_5) = [4,6]$.
The interval $[4,6]$  is a violated Hall interval as it contains
four variables. We show that Equations~\eqref{eq:max_set_1}--\eqref{eq:max_set_2}
detect that the interval $[3,6]$ has to be pruned from $D(X_1)$.

Consider the pair $X_1=3$ and the interval $[v,v+k]$,
where $v=3$, $k=3$. We get that 
$B^1_{1,{6}}   =  |\{j \in \{2,3\} |  \myset{D}(X_j) \subseteq [1, 6]\}| =2$
$D^1_{3,6}  = |\{j \in \{4,5\}) | \myset{D}(X_j) \subseteq [3, 6]\}| = 2$
and $B^1_{1,{6}} + D^1_{3,6} = 4$ which is greater than $k =3$.
Hence, the interval $[3,6]$ has to be removed from  $D(X_1)$.
\end{myexample}

\begin{theorem}
\label{t:bound_consistency_conditions}
Consider the $\alldiff[X_1,\ldots,X_n]$ constraint and a set of precedence constraints $X_i < X_j$.
Enforcing conditions~\eqref{eq:max_set_1}--\eqref{eq:min_sum} together with bounds consistency
on the $\alldiff$ constraint and the precedence constraints is equivalent to
enforcing bounds consistency on  the $\alldiffprec$ constraint.
\end{theorem}
\myprroof
Suppose conditions~\eqref{eq:max_set_1}--\eqref{eq:min_sum} are fulfilled,
$\alldiff$ and precedence constraints are bounds consistent
and the $\alldiffprec$ constraint is not bounds consistent.
Let an assignment of a variable $X_i$ to its bound $\max(X_i)$
be an unsupported bound. \nina{We denote $\max(X_i)$  $v$ to simplify notations.}
We recall that we denoted the domains after direct pruning of $X_i=v$
$\Ddpv(X_1),\ldots, \Ddpv(X_n)$.
 By Lemma~\ref{l:testing bound}
the  $\alldiff([X_1,\ldots, X_n])$ constraint where
domains of variables $X$ are $\Ddpv(X_1),\ldots, \Ddpv(X_n)$ fails.
Hence,  there exists a violated Hall interval $[l,u]$ such that
$|\Ddpv(X_i) \subseteq [l,u] \}| > u -l +1$.

Note that direct pruning of $X_i = v$ does not cause
the pruning of  variables in $P(i)$, as all precedence constraints
are bounds consistent on the original domains.
Next we consider several cases depending on the relative position
of the value $v$ and the violated Hall interval on the line.
Note that the interval $[l,u]$ was not a violated Hall interval
before the assignment $X_i=v$. However, due to direct pruning of $X_i=v$
a number of additional variables domains can be forced to be
inside $[l,u]$. Hence, we analyze these additional variables
and show that conditions~\eqref{eq:max_set_1}--\eqref{eq:min_sum} prevent the creation of a violated Hall interval.

\noindent \textbf{Case 1.} Suppose  $v \in [l,u]$. As $[l,u]$ is a violated Hall
  interval, we have that
$$|\{j \in S(i) | \Ddpv(X_j) \subseteq [l,u]]\}|  + |\{j \notin S(i) |\Ddpv(X_j) \subseteq [l, u]\}| > u - l,$$

\noindent Note that the number of additional variables
that fall into the interval $[l,u]$ after setting  $X_i$ to $v$
consists only of variables that succeed $X_i$, such that
$\myset{D}(X_j) \subseteq [1,u]$. Hence,
$|\{j \notin S(i) | \Ddpv(X_j) \subseteq [l, u]\}| = |\{j \notin S(i) | \myset{D}(X_j) \subseteq [l, u]\}|$,
$|\{j \notin S(i) | \Ddpv(X_j) \subseteq [l, u]\}| = |\{j \in S(i) | \myset{D}(X_j) \subseteq [1, u]\}|$ and

$$|\{j \in S(i) | \myset{D}(X_j) \subseteq [1, u]\}|  +
|\{j \notin S(i) | \myset{D}(X_j) \subseteq [l, u]\}| > u - l,$$
which violate conditions~\eqref{eq:max_set_1}--\eqref{eq:max_sum}
for $v = l$ and $k = u - l$.

\noindent  \textbf{Case 2.} Suppose  $v \notin [l,u]$. If $v > u + 1$
or $v < l - 1$, the assignment $X_i = v$ does not force
any extra variables to fall into the interval $[l,u]$. \nina{Hence,
the interval $[l,u]$ is a violated Hall interval before
the assignment. This contradicts that $\alldiff$ is bounds consistent.}

\noindent  \textbf{Case 3.}  Suppose $v = u + 1$. In this case
the assignment  $X_i = v$  does not force any additional variables among successors
to fall into $[l,u]$, as $\Ddpv(X_j) \subseteq [u+2,d]$.
\nina{Note that there are no successors that are contained in the interval $[1,v]$,
because precedence constraints are bounds consistent. Therefore,
$|\{j \in S(i) | \myset{D}(X_j) \subseteq [l, v]\}| = 0$.
Hence, the only additional variables that fall into $[l,u]$ are variables
that do not have a precedence relation with $X_i$ and $v = \max(X_j) = u + 1$,
so $|\{j| j \notin S(i),\Ddpv(X_j) \subseteq [l,u]\}| = |\{j| j \notin S(i), \myset{D}(X_j) \subseteq 
[l,u+1]\}|$. As $[l,u]$ is a violated Hall interval, we have
$$|\{j| j \notin S(i), \myset{D}(X_j) \subseteq
[l,u+1]\}| =|\{j| j \notin S(i),\Ddpv(X_j) \subseteq [l,u]\}| > u - l + 1.$$
This contradicts Equation~\eqref{eq:min_sum}
$|\{j \in S(i) | \myset{D}(X_j) \subseteq [l, u+1]\}|+ |\{j| j \notin S(i), \myset{D}(X_j) \subseteq
[l,u+1]\}| \leq (u +1) -l$ 
as the first term equals 0 in the equation by the argument above.}

\noindent \textbf{Case 4.} Suppose $v = l - 1$. In this case the set of additional variables
that fall into the interval $[l,u]$ consists of two subsets of variables.
The first set contains variables that succeed $X_i$, such that $\myset{D}(X_j) \subseteq [l',u]$, $l' < v$
and $\Ddpv(X_j) \subseteq [l,u]$. The second set contains the variables
that do not have precedence relation with $X_i$ and $v =max(X_j) = l - 1$.
Consider the interval $[l-1, u]$.
As conditions~\eqref{eq:max_set_1}--\eqref{eq:max_sum} are satisfied for the interval $[l-1,u]$,
we get that
$$|\{j \in S(i) |  \myset{D} (X_j) \subseteq [1,u]\}|  +
|\{j \notin S(i) | \myset{D}(X_j) \subseteq [l-1, u]\}| \leq u - (l-1),$$
\noindent  On the other hand, as the $[l,u]$ is violated we have
$$|\{j \in S(i) |   \Ddpv(X_j)  \subseteq [l,u]\}|  +
|\{j \notin S(i) | \Ddpv(X_j) \subseteq [l, u]\}| > u - l + 1,$$
We know that
$|\{j \notin S(i) | \myset{D}(X_j) \subseteq [l-1, u]\}| = |\{j| j \notin S(i), \Ddpv(X_j) \subseteq [l,u]\}| $ and $|\{j \in S(i) |  \myset{D} (X_j) \subseteq [1,u]\}| = |\{j \in S(i) |   \Ddpv(X_j)  \subseteq [l,u]\}|$ by the construction of the direct pruning. This leads to a contradiction between the last two inequalities.

Therefore, the interval $[l,u]$ cannot be a violated Hall interval.
Similarly, we can prove the same result for the minimum value of $X_j$.

The reverse direction is trivial.
\qed

Theorem~\ref{t:bound_consistency_conditions} proves that conditions~\eqref{eq:max_set_1}--\eqref{eq:min_sum}
together with bounds consistency on the $\alldiff$ constraint and the precedence constraints
are necessary and sufficient conditions  to enforce bounds consistency on the
$\alldiffprec$ constraint. The time complexity of enforcing these conditions
in $O(nd^2)$, as for each variable we check $O(d^2)$ intervals.
This time complexity can be reduced by making an observation,
that we do not need to check  intervals of length greater than $n$
as conditions are trivially satisfied for such intervals. This reduces the complexity
to $O(n^2d)$.

We make an observation that helps to further reduce the time complexity of enforcing these conditions.
We denote $L$ the set of all minimum values in variables domains $ L = \cup_{i=1}^n \{\min(\myset{D}(X_i))\}$
and $U$ the set of all maximum values in variables domains $U = \cup_{i=1}^n \{\max(\myset{D}(X_i))\}$.
Let $[l,u]$ be an interval that violates the conditions. We denote $c_{l,u}$ the
amount of violation in this interval:
$c_{l,u} = B^i_{1,u}  +
D^i_{l,u} - (u-l).$


\begin{myobservation}
\label{o:lool_at_bounds_only}
Let $X_i$ be a variable and $[v,v+k]$, $v \in \myset{D}(X_i)$ be an interval that violates
 conditions~\eqref{eq:max_set_1}--\eqref{eq:max_sum}. Then there exists a violated interval $[l,u]$ such that
$[l,u] \subseteq [v,v+k]$, $l,u \in L \cup U$ and $c_{l,u} > l - v$.
\end{myobservation}
\myprroof
Consider a violated interval $[v,v+k]$. In this case $B^i_{1,v+k}  +  D^i_{v,v+k} > k$.
There exists an interval $[l,u] \subseteq [v,v+k]$ such that $l,u \in L \cup U $ . We take the largest interval $[l,u]$.
Note that such an interval always exists as the interval $[\max(X_i), \max(X_i)]$ is contained inside the interval $[v,v+k]$.
The interval $[l,u]$ also
violates the conditions, because it contains the same variables. So, we have $B^i_{1,u}  +  D^i_{l,u} > u-l$.
We note that $D^i_{l,u} = D^i_{v,v+k}$
as there are no lower bounds in the interval $[v,l)$.
Similarly, there are no upper bounds in the interval $(u,v+k]$. Hence, $B^i_{1,u} = B^i_{1,v+k}$.
Therefore, $B^i_{1,u}  + D^i_{l,u} > k$.
The value $c_{l,u}$ is greater than $k - u + l \geq v + k - v - u + l \geq  v + k - u + l - v \geq l - v$
as $ u \leq  v + k$.


\qed
Observation~\ref{o:lool_at_bounds_only} shows that it is sufficient
to check intervals $[v, v + k]$, 
$\{v, v + k\} \in L \cup U$.
We can infer all pruning from these intervals. Let $[l,u]$, $l, u \in L \cup U$ be an interval that violates conditions~\eqref{eq:max_set_1}--\eqref{eq:max_sum} for a variable $X_i$ and $c_{l,u}$ be the violation cost.
Then we remove the interval $[l - (c_{l,u} -1), u]$ from $\myset{D}(X_i)$, as any interval between
$[l - (c_{l,u} -1), u]$ and $[l, u]$ is a violated interval.
A dual observation holds for conditions~\eqref{eq:min_set_1}--\eqref{eq:min_sum}.
This reduces the time complexity of checking~\eqref{eq:max_set_1}--\eqref{eq:min_sum} to $O(n^3)$.

\section{Faster bounds consistency algorithm}

\newcommand{\dom}{\myset{D}}
\begin{algorithm}[t]
\label{a:faster_bc}
  \caption{PruneUpperBounds($X_1, \ldots, X_n$)}
  Sort variables such that $\max(\dom(X_i)) \leq \max(\dom(X_{i+1}))$\;
  \For{$i \in 1..n$}{ \label{a:for_loop}
    Create a disjoint set data structure $T$ with the integers $1..d$\;
    $b \gets \max(\dom(X_1))+1$\;
    Invariant: $b$ is the smallest value such that there are exactly
    as many available values in the open-interval $[b,
    \max(\dom(X_j))+1)$ as there are successors of $X_i$ that have been processed.\;
    \For{$X_j$ in non-decreasing order of upper bound}{
      \If{$j \not\in S(i)$}{
        $S \gets \mbox{Find}(\min(\dom(X_j)), T)$\;
        $v \gets \min(S)$\; \label{a:min}
        Union($v, \max(S) + 1$, T)\; \label{a:union}
      }
       \If{$j > 1$}{
        \For{$k \in 1..\max(\dom(X_j)) - \max(\dom(X_{j-1}))$}{
          $b \gets \max(Find(b, T)) + 1$\; \label{a:b_forward}
        }
        \If{$Find(v, T) = Find(b, T) \lor v > b \lor j \in S(i)$}{
          $b \gets \min(Find(b - 1, T))$\; \label{a:b_backward}
        }
      }
        $\max(\dom(X_i)) \gets \min(\max(\dom(X_i)), b - 1)$\;
    }
  }
\end{algorithm}

Observation~\ref{o:lool_at_bounds_only} allows us to construct a faster algorithm to enforce  conditions~\eqref{eq:max_set_1}--\eqref{eq:min_sum}.  First,
we observe that the conditions can be checked for each variable independently.
Consider a variable $X_i$. We sort all variables  $X_j$, $j=1,\ldots,n$ in a non-decreasing order of their upper bounds.
When processing a variable $X_j$, $j \notin S(i)$,
we assign $X_j$ to the smallest value that has not been taken.
When processing a variable $X_j$, $j \in S(i)$,
we store  information about the number of successors that we have seen so far.
We perform pruning if we find an interval $[l,u]$ such that
the number of available values in this interval  equals
the number of successors in the interval $[1,u]$.
We use a disjoint set data structure to perform counting operations
in $O(d)$ time.

Algorithm~\ref{a:faster_bc} shows a pseudocode of our algorithm.
\nina{We denote $T$ a disjoint set data structure.}
The function $Find(v_1,T)$ returns the set that contains the value $v_1$.
The function $Union(v_1,v_2, T)$ joins the values $v_1$ and $v_2$ into a single set.
We use a disjoint set union data structure~\cite{Gabow83} that allows to
perform $Find$ and $Union$ in $O(1)$ time.
\begin{mytheorem}
\label{t:faster_bc}
Algorithm~\ref{a:faster_bc} enforces conditions~\eqref{eq:max_set_1}--\eqref{eq:max_sum}
in $O(nd)$ time.
\end{mytheorem}
\myprroof
Enforcing conditions~\eqref{eq:max_set_1}--\eqref{eq:max_sum} on
the $i$th variable corresponds to the $i$th loop (line~\ref{a:for_loop}).
Hence, we can consider each run independently.

We denote $I_j$ a set of values that are taken by non-successors of $X_i$
after the variable $X_j$ is processed. The algorithm maintains a pointer $b$ that
stores the minimum value such that the number of available values
in the interval $[b,\max(X_j) + 1)$ is equal to $B^i_{1,max(X_j)}$
after the variable $X_j$ is processed.

\textbf{Invariant}.
We prove the invariant for the pointer $b$ by induction.
The invariant holds at step $j = 0$.
Note that the first variable can not be a successor of $X_i$.
Indeed, $b = max(X_1) +1$ and the interval $[\max(X_1) +1,\max(X_1) +1)$
is empty. Let us assume that the invariant holds after processing the variable $X_{j-1}$.

Suppose the next variable to process is $X_j$. After we assigned $X_j$ to a value,
we move 
$b$ forward to capture a possible
increase of the upper bound from $\max(X_{j-1})$ to $\max(X_j)$ (line~\ref{a:b_forward}) and, then,
backward if either $X_j$ is a successor of $X_i$ or  $X_j$ is a non-successor and $X_j$ takes a value $v$ such that
$b \leq v$ (line~\ref{a:b_backward}).
Note, that when we move $b$, we ignore  values in $I_j$.
To point this out we call steps of $b$ available-value-steps.
Thanks to a disjoint set union data structure we can jump over values in $I_j$ in $O(1)$ per step~\cite{Gabow83}.

\textbf{Moving forward}. We move the pointer $b$ on $\max(X_j) - \max(X_{j-1})$ available-value-steps forward.
We denote  $b'$ a new value of $b$.  The line~\ref{a:b_forward} ensures that the number of available values
in the interval $[b',\max(X_j) + 1)$ equals to
the number of available values  in the interval $[b,\max(X_{j-1}) + 1)$.
This operation preserves the invariant by the induction hypothesis.

\textbf{Moving backward}.
We consider two cases.

\textbf{Case 1.} $X_j$ is a successor of $X_i$. In this case,
we move $b'$ one available-value-step backward to capture that $X_j$ is a successor (line~\ref{a:b_backward}).
This preserves the invariant.

\textbf{Case 2.} $X_j$ is not a successor of $X_i$.
Suppose $v$ and $b'$ are in the same set, so that $Find(v,T) =  Find(b,T)$. Then we
move $b'$ to the minimum element in this set. This step does not change
the number of available values between the pointer $b'$ and $\max(X_{j})$.
However, it makes sure that $b'$ stores the minimum possible
value. This preserves the invariant.

Suppose $v$ and $b'$ are in different sets.
If $v > b'$ then we  move $b'$ one available-value-step backward,
as $v$ took one of the available values in $[b',\max(X_j) + 1)$.
This preserves the invariant.
If $v < b'$ then the invariant holds by the induction hypothesis. 
Hence, the new value of $b$ preserves the invariant.

Note that the length of the interval $[b,\max(X_j) +1)$  equals  the sum of $B^i_{1,\max(X_j)}$
and $D_{b,\max(X_j)}$ due to the invariant. This means that the interval $[b,\max(X_j) +1)$
violates conditions~\eqref{eq:max_set_1}--\eqref{eq:max_sum}, as the sum  $B^i_{1,\max(X_j)} + D_{b,\max(X_j)}$
has to be  less than or equal to the length of the interval $[b,\max(X_j) + 1)$ minus 1.

\textbf{Soundness}.
Suppose we pruned an interval  $[b-1,\max(X_j)]$ from $\myset{D}(X_i)$ after the processing of the variable $X_j$.
This pruning is sound because the  interval $[b,\max(X_j) +1)$  violates conditions~\eqref{eq:max_set_1}--\eqref{eq:max_sum}.

\textbf{Completeness}.
Suppose there exists an interval $[l,u]$ that violates conditions~\eqref{eq:max_set_1}--\eqref{eq:max_sum}, so that
$B^i_{1,u}  +  D^i_{l,u} > u-l$. However, the algorithm does not prune the upper bound of $X_i$ to $l-1$.
Suppose that $l \in L$, $u \in U$. As the pointer $b$ preserves the invariant,
there are exactly $B^i_{1,u}$ available values between $[l, u+1)$.
Hence $b$ points to $l$ and  $\max(X_i) \leq l-1$.

Suppose that $l \notin L$, $u \in U$. We consider the step when the last pruning of the variable $X_i$
occurs. Suppose we processed the variable $X_j$ at this step.
The pointer $b$ stores $\max(X_i)+1$. As $b$ does not move backward in the following steps,
we conclude that neither successors nor non-successors with domains that are contained inside the interval $[b,d]$
occur. Hence, $B^i_{1,u}  +  D^i_{l,u} = B^i_{1,u}  +  D^i_{\max(X_i)+1,u}$, $\max(X_j) \leq u$, $u \in U$, $l < \max(X_i)$.
Hence $[l,u]$ is not a violated interval.

\textbf{Complexity}. At each iteration of the loop (line~\ref{a:for_loop}) the pointer $b$ moves $O(d)$ times forward and $O(n)$ times backward.
Due to a disjoint set data structure the total cost of the  operations is $O(d)$, the functions
$Union(v_1, v_2, T)$ and  $Find(v_1, T)$  take $O(1)$~\cite{Gabow83}. The total time
complexity is $O(nd)$.
\qed

We can construct a similar algorithm to Algorithm~\ref{t:faster_bc} to enforce conditions~\eqref{eq:min_set_1}--\eqref{eq:min_sum} and prune lower bounds.

\begin{figure}[t]
\centering
\label{f:algo_exm}
\includegraphics[width=0.85\textwidth]{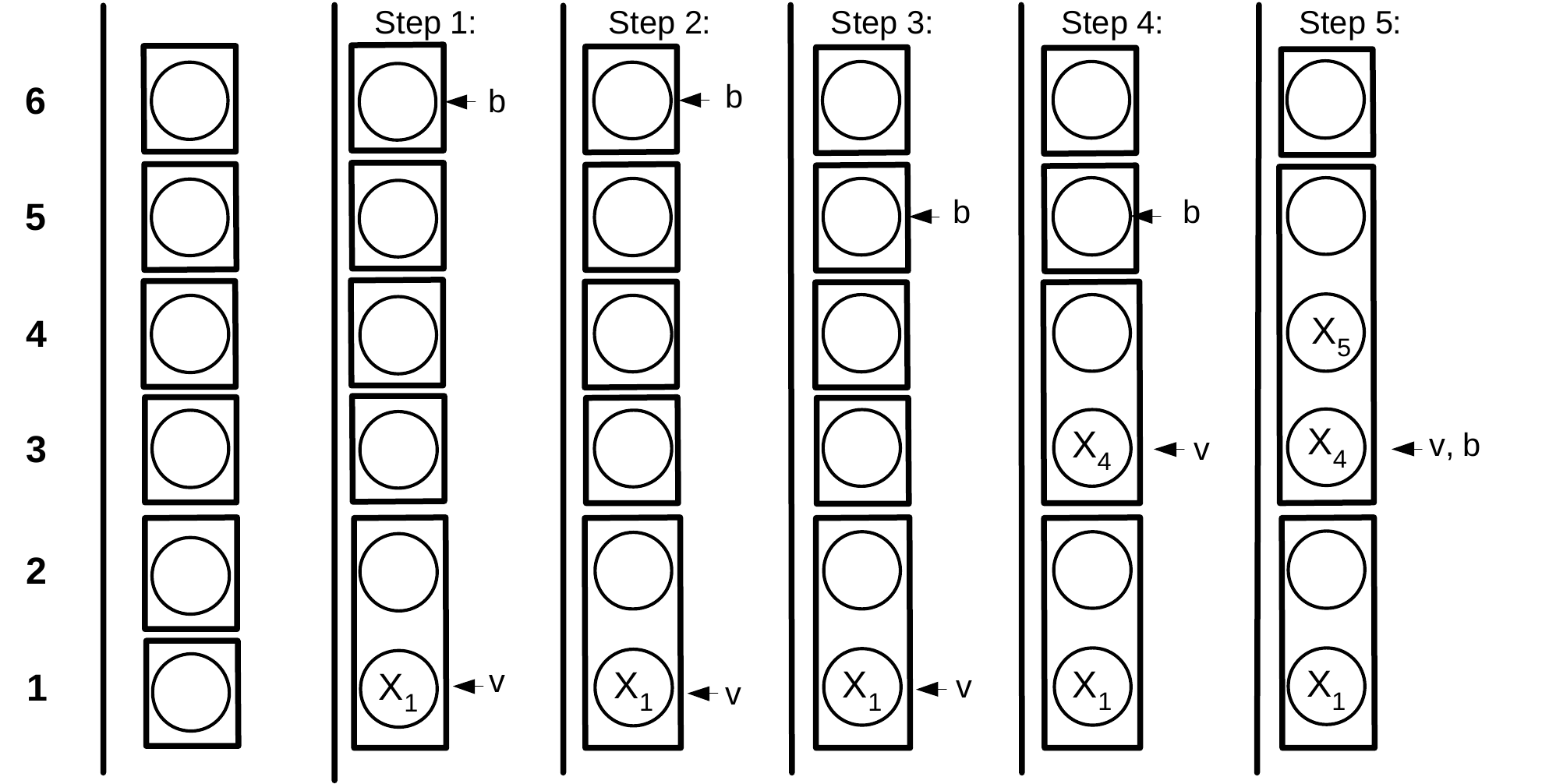}
\caption{Algorithm~\ref{a:faster_bc} enforces conditions~\eqref{eq:max_set_1}--\eqref{eq:max_sum} on the variable $X_1$.}
\end{figure}
\begin{myexample}
\label{exm:faster_bc_1}
Consider $\alldiffprec ([X_1,X_2,X_3,X_4,X_5],\{(1,2),(1,3)\})$
for Example~\ref{exm:faster_bc}. We show how our algorithm works
on this example.

We represent values in the disjoint set data structure $T$ with
circles. We use rectangles to denote
sets of joint values. Initially, all values are in disjoint sets.
If a variable $X_i$ takes a value $v$ we put the label
$X_i$ in the $v$th circle.
Figure~\ref{f:algo_exm} shows five steps of the algorithm
when processing the variable $X_1$ (line~\ref{a:for_loop}, $i=1$).

Consider the first step. We set $v=1$
as $\min(X_1)$ is 1. We join the values $1$ and $2$
into a single set (line~\ref{a:union}). The pointer $b$ is
set to $\max(X_1) + 1 = 6$.
Consider the second step. We process the variable $X_2$ which is a successor
of $X_1$. As  $\max(X_2) - \max(X_1) = 1$ we move $b$ one
available-value-step forward, $b = 7$. However, as $X_2$ is a successor,
we move $b$ available-value-step backward. Hence, $b = 6$.
Consider the third step. We process 
$X_3$ which is a successor
of $X_1$. As  $\max(X_3) - \max(X_2) = 0$ we do not move $b$ forward.
However, as $X_3$ is a successor, we move $b$ available-value-step backward,
$b$ is set to 5.
Consider the fourth step. We process 
$X_4$ which is a non-successor
of $X_1$. The value $\min(X_4)$ is 3. Hence, $v = 3$
and join $3$ and $4$ into a single set.
Consider the fifth step. We process the variable $X_5$ which is a non-successor
of $X_1$. The value $\min(X_5)$ is 4, as the value $3$ is taken by $X_4$. As values $3$ and $4$
are in the same set, we do not move $v$ and join $\{3,4\}$ and $5$ into a  set.
Note that $v$ and $b$ are in the same set and we move $b$ to the minimum
element in this set. Hence, $b = 3$ and we prune $[3,5]$ from $X_1$.
\end{myexample}

The complexity of the algorithm can be reduced to $O(n^2)$. Let $L$ be
the set of domain lower bounds sorted in increasing order and let
$l_{i-1}$ and $l_{i}$ be two consecutive values in that
ordering. Following~\cite{lopez1}, we initialize the disjoint set data
structures with only the elements in $L$. We assign a counter $c_i$ to
each element $l_i$ initialized to the value $l_{i} - l_{i-1}$.
Line~\ref{a:union} of the algorithm can be modified to decrement the
counter of $\max(S)$. The algorithm calls the function \emph{Union}
only if the counter of $\max(S)$ is decremented to zero. The algorithm
preserves its correctness and since there are at most $n$ elements in
$L$, the factor $d$ in the complexity of the algorithm is replaced by
$n$ resulting in a running time complexity of $O(n^2)$.

\section{Bounds consistency decomposition}

We present a decomposition of the $\alldiffprec$ constraint.
For $1 \leq i \leq n$, $1 \leq l \leq u \leq d$ and $u-l < n$,
we introduce Boolean variables $B_{il}$ and $A_{ilu}$ 
and post the following constraints:
\begin{eqnarray}
B_{il}=1 & \iff & X_i \leq l \label{eq:decomposition_channel_1} \\
A_{ilu}= 1 & \iff & (B_{i(l-1)}=0 \wedge B_{iu}=1) \label{eq:decomposition_channel_2}\\
\sum_{i=1}^n A_{ilu} & \leq & u - l + 1 \label{eq:decomposition_intervals}\\
\sum_{j \in S(i)} A_{j,1,u}  + \sum_{j \notin S(i)} A_{j,l,u} - B_{i(l-1)} &\leq& u - l \label{eq:decomposition_suc}\\
\sum_{j \in P(i)} A_{j,l,d}  + \sum_{j \notin P(i)} A_{j,l,u} - (1 - B_{iu}) &\leq& u - l\label{eq:decomposition_pred} \\
\forall j \in S(i), X_i &<& X_j  \label{eq:decomposition_less_1} \\
\forall j \in P(i), X_{j} &<& X_i \label{eq:decomposition_less_2} 
\end{eqnarray}

\begin{theorem}
  \label{t:bound_consistency_decomposition}
  Enforcing bounds consistency
  on constraints \eqref{eq:decomposition_channel_1}
  and \eqref{eq:decomposition_less_2} enforces bounds consistency on the
  corresponding $\alldiffprec$ constraint in $O(n^2d^2)$ down a
  branch of the search tree.
\end{theorem}
\myprroof
Constraints~\eqref{eq:decomposition_channel_1}--\eqref{eq:decomposition_intervals}
enforce bounds consistency on the $\alldiff$ constraint.
Constraints~\eqref{eq:decomposition_less_1}--\eqref{eq:decomposition_less_2} enforce bounds consistency
on the precedence constraints.
Finally, conditions~\eqref{eq:min_set_1}--\eqref{eq:min_sum}
are captured by constraints~\eqref{eq:decomposition_suc} and  \eqref{eq:decomposition_pred}.
By Theorem~\ref{t:bound_consistency_conditions}, enforcing BC on $\alldiff$,
precedence constraints and enforcing conditions~\eqref{eq:min_set_1}--\eqref{eq:min_sum} is sufficient to
enforce bounds consistency on the $\alldiffprec$ constraint.
The time complexity is dominated by $O(nd^2)$ linear inequality constraints~\eqref{eq:decomposition_suc}--\eqref{eq:decomposition_pred}. It takes $O(n)$ time
to propagate a linear inequality constraint over $O(n)$ Boolean variables
down a branch of the search tree.
Hence, the total complexity is  $O(n^2d^2)$.
\myqed

Note that the time complexity of decomposition contains a factor $d$ that we cannot
reduce as in the case of the conditions~\eqref{eq:max_set_1}--\eqref{eq:min_sum}.
As we compute the time complexity down a branch of a search tree we have to consider all
possible $O(d^2)$ tight intervals that might emerge during the search.

%
%

\section{Domain consistency}

Whilst enforcing bounds consistency
on the \alldiffprec constraint
takes just low order polynomial time,
enforcing domain consistency is intractable
in general (assuming $P \neq NP$).

\begin{mytheorem}
Enforcing domain consistency
on $\alldiffprec([X_1,\ldots,X_n],E)$
is NP-hard.
\end{mytheorem}
\myprroof
We give a reduction from 3-SAT.
Suppose we have a 3-SAT problem in $N$ variables
and $M$ clauses.
We consider an $\alldiffprec$ constraint on $2N+3M$ variables.
The first $2N$ variables represent a truth assignment.
The next $3M$ variables represent the literals
which satisfy each of the clauses.
For $1 \leq i \leq N$,
the variables $X_{2i-1}$ and
$X_{2i}$ have domains $\{i,N+M+i\}$.
$X_{2i-1}=i$ corresponds to the
case in which we have a truth assignment that assigns $x_i$ to false
whilst $X_{2i}=i$ corresponds to the
case in which we have a truth assignment that assigns $x_i$ to true.
The all different constraint ensures that
only one of $X_{2i-1}$ and $X_{2i}$ can be
assigned to $i$. Hence one of these two cases must hold.
For $1 \leq i \leq M$,
the variables $X_{N+3i-2}$, $X_{N+3i-1}$ and
$X_{N+3i}$ represent the three literals in
each clause. The values assigned to these variables
will ensure that the truth assignment
satisfies at least one literal in each clause.
The domains of $X_{N+3i-2}$, $X_{N+3i-1}$ and
$X_{N+3i}$ are $\{N+i,2N+M+2i,2N+M+2i-1,\}$.
$N+i$ will be the value used to indicate that the
corresponding literal satisfies the clause.
For each literal in a clause, we add
an edge to $E$ to ensure that there is
an ordering constraint between one of the
first $2N$ variables in the truth assignment
section and the corresponding
variable in the clause section.
For example, suppose the $i$th clause is
$x_j \vee \neg x_k \vee x_l$ then
we add 3 edges to $E$ to
ensure: $X_{2j}<X_{N+3i-2}$,
$X_{2k-1}<X_{N+3i-1}$,
and $X_{2l}<X_{N+3i}$.
The all different constraint ensures
one of $X_{N+3i-2}$, $X_{N+3i-1}$ and
$X_{N+3i}$ takes the smallest value
$N+i$, and the ordering constraint
then checks that the corresponding
literal is set to true. By construction,
the $\alldiffprec$ constraint has
support iff there is a satisfying
assignment to the original 3-SAT problem.
\myqed

Note that the proof uses a DAG defined by $E$
that is flat, and does not contain any chains. Hence,
enforcing domain consistency on $\alldiffprec$
remains NP-hard without chains of precedences.
Note also that SAT remains NP-hard even if
each clause has at most 3 literals, and each
literal or negated literal occurs at most
three times. Hence, a similar reduction shows
that enforcing domain consistency on $\alldiffprec$
remains NP-hard even if the degree of nodes
in $E$ is at most 3 (that is, we have at most
3 precedence constraints on any variable).

\section{Other related work}

There have been many studies on propagation algorithms for a single
\alldiff constraint.  A domain consistency algorithm that runs in
$O(n^{2.5})$ was introduced in~\cite{regin1}. A range consistency
algorithm was then proposed in~\cite{Leconte} that runs in time
$O(n^2)$. The focus was moved from range consistency to bound
consistency with~\cite{puget98}, who proposed a bounds consistency
algorithm that runs in $O(n\log n)$. This was later improved further
in~\cite{mtcp02} and then in~\cite{lopez1}.

Decompositions
that achieve bounds consistency have
been given for a number of global
constraints. Relevant to this work,
similar decompositions have been
given for
a single \alldiff constraint \cite{bknqwijcai09},
as well as for
overlapping \alldiff\ constraints
\cite{bknqwaaai2010}.
These decompositions have
the property that enforcing bound
consistency on the decomposition
achieves bounds consistency on
the original global constraint.

A number of global constraints have been
combined together and specialized propagators
developed to deal with these conjunctions.
For example, a global lexicographical ordering and sum
constraint have been combined together \cite{hkwaimath04}.
As a second example, a generic method has been proposed
for propagating combinations of the global lexicographical ordering and
a family of globals including the {\sc Regular} and {\sc Sequence}
constraints \cite{knwercim09}.

\section{Conclusions}

We have proposed a new global constraint that combines together an \alldiff
constraint with precedence constraints that strictly order given pairs
of variables.
We
gave an efficient propagation algorithm that enforces bounds consistency
on this global
constraint in $O(n^2)$ time,
and showed how this propagator can be simulated with a
simple decomposition
extends the bounds consistency enforcing decomposition proposed for the
\alldiff constraint.
Finally, we proved that enforcing domain
consistency on this global constraint is NP-hard in general.
There are many interesting future directions. We could, for example,
study the convex hull of the \alldiffprec constraint. Other interesting future work
includes studying the combination of
precedence constraints with generalizations of the \alldiff constraint
including the global cardinality constraint and the inter-distance
constraint.

\bibliographystyle{splncs}

\bibliography{/Users/twalsh/Documents/biblio/a-z,/Users/twalsh/Documents/biblio/pub,/Users/twalsh/Documents/biblio/a-z2,/Users/twalsh/Documents/biblio/pub2}

\end{document}